\documentclass[runningheads]{llncs}
\usepackage{graphicx}
\usepackage{cite}
\usepackage{amsmath,amssymb,amsfonts}
\usepackage{algorithmic}
\usepackage{graphicx}
\usepackage{xcolor}
\usepackage{booktabs} 
\usepackage{multirow} 
\usepackage{marvosym} 
\begin{document}
\title{FracDetNet: Advanced Fracture Detection via Dual-Focus Attention and Multi-scale Calibration in Medical X-ray Imaging
}
\titlerunning{FracDetNet: Dual-Focus and Multi-scale Fracture Detection}
%
%


\author{Yuyang Sun\inst{1} \and
Cuiming Zou\inst{1}\textsuperscript{(\Letter)}}

\authorrunning{Y. Sun and C. Zou}

\author{Yuyang Sun\inst{1} \and
Cuiming Zou\inst{1}\textsuperscript{(\Letter)}}
\institute{Huazhong Agricultural University, Wuhan, China \\
\email{frontsea040320@gmail.com, zoucuiming2006@163.com}\\
\textsuperscript{(\Letter)}Corresponding author}

\maketitle              
\begin{abstract}
In this paper, an advanced fracture detection framework, FracDetNet, is proposed to address challenges in medical imaging, as accurate fracture detection is essential for enhancing diagnostic efficiency in clinical practice. Despite recent advancements, existing methods still struggle with detecting subtle and morphologically diverse fractures due to variable imaging angles and suboptimal image quality. To overcome these limitations, FracDetNet integrates Dual-Focus Attention (DFA) and Multi-scale Calibration (MC). Specifically, the DFA module effectively captures detailed local features and comprehensive global context through combined global and local attention mechanisms. Additionally, the MC adaptively refines feature representations to enhance detection performance. Experimental evaluations on the publicly available GRAZPEDWRI-DX dataset demonstrate state-of-the-art performance, with FracDetNet achieving a mAP$_{50-95}$ of 40.0\%, reflecting a \textbf{7.5\%} improvement over the baseline model. Furthermore, the mAP$_{50}$ reaches 63.9\%, representing an increase of \textbf{4.2\%}, with fracture-specific detection accuracy also enhanced by \textbf{2.9\%}. The code will be made publicly accessible once accepted.

\end{abstract}
\section{Introduction}

\indent As artificial intelligence (AI) continues to advance rapidly, computer vision technology has been increasingly applied in the medical field, particularly in medical imaging analysis \cite{hardalac2022wrist, kuo2022ai, yao2021rib, chien2024yolov8am}. One of the most significant research directions is the utilization of object detection techniques to identify fracture regions in X-ray images, this approach has been extensively adopted in clinical settings.

Pediatric fractures have emerged as a pressing public health concern in modern society, necessitating efficient and accurate diagnostic techniques. According to the study by Naranje et al. \cite{naranje2016epidemiology}, fractures account for a significant portion of pediatric emergency cases in the United States, with children aged 10 to 14 years being at the highest risk. However, traditional fracture diagnosis heavily relies on the expertise of medical professionals, demanding extensive cognitive skills and technical proficiency. The presence of subtle fractures and morphologic changes complicates the diagnostic process and often leads to prolonged evaluation times, a challenge that is particularly acute in areas with limited medical resources.

There has been notable progress in applying deep learning-based object detection techniques for identifying fractures from X-ray images \cite{chien2024yolov9}. The integration of artificial intelligence-assisted detection not only helps clinicians make timely and accurate diagnoses but also reduces human errors in fracture assessment, which is particularly important in resource-limited environments.
\begin{figure}[h]
	\centering
	\includegraphics[width=0.8\columnwidth]{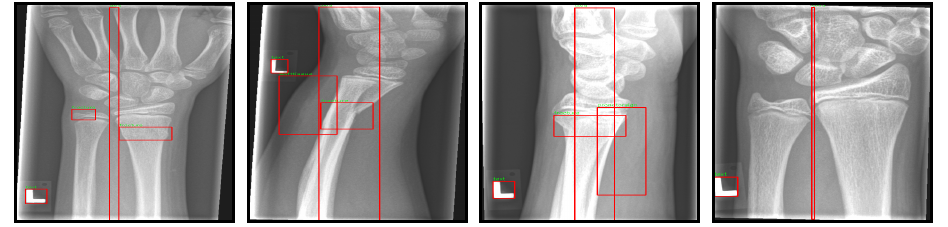}
	\caption{Labeled X-ray image samples.}
	\label{fig1}
\end{figure}
Nevertheless, previous studies have identified three primary challenges in the clinical implementation of fracture detection \cite{joshi2020survey, meena2022bone, langerhuizen2019applications}: as shown in Fig.~\ref{fig1}, limited image quality has been reported to hinder the effective detection of subtle fractures, large variations in fracture morphology increase diagnostic complexity, and diverse imaging angles resulting from varying patient postures introduce perspective effects that complicate classification and localization tasks.

To address the real-time demands and overcome the constraints of limited computational resources in pediatric fracture detection \cite{hardalac2022wrist}, we sought to improve existing object detection models and ultimately adopted YOLOv8 as the foundational detection framework. YOLOv8 is selected for its mature and stable performance, particularly excelling in small object detection, making it highly suitable for medical imaging tasks. Building upon this, an enhanced attention mechanism Dual-Focus Attention (DFA) is introduced \cite{Shi_2024_CVPR}, which leverages global as well as local cues to strengthen the model’s capacity for capturing fracture-related features within intricate medical imagery, effectively mitigating the impact of perspective distortion and poor image quality. Furthermore, inspired by multi-scale perception structures \cite{huang2024channel}, the Multi-scale Calibration Head (MC) is redesigned, which further improves the model's robustness in detecting fractures at various scales while attaining an optimal trade-off between computational efficiency and accuracy. Based on these improvements, a model particularly optimized for pediatric fracture recognition, named Fracture Detection Network (FracDetNet), is developed.

Our main contributions are concisely presented as below:
\begin{itemize}
	\item The model first introduces an improved Dual-Focus Attention (DFA) mechanism for fracture detection. This mechanism enhances self-attention by leveraging both global and local contextual information to attenuate the influence of perspective distortion and poor image quality on fracture detection.
	
	\item An improved Multi-scale Calibration Head (MC) is proposed, which enhances the detector's multi-scale recognition ability, thereby strengthening the model's robustness. This represents an innovative attempt to address the multi-scale issue in fracture detection from a novel perspective.
	
	\item Based on the modifications to DFA and MC, Fracture Detection Network (FracDetNet) is proposed. It effectively balances high accuracy with low computational cost. Comprehensive experiments were carried out on publicly available datasets have shown that FracDetNet attains state-of-the-art (SOTA) performance, highlighting its clinical potential.
\end{itemize}

\section{Related Works}

Deep learning has significantly improved fracture detection, enhancing both accuracy and efficiency. Recent advancements can be categorized into model development and attention mechanisms.

In model development, Chien \textit{et al.} \cite{chien2024yolov9} introduced YOLOv9 for pediatric wrist fracture detection, achieving a mAP$_{50-95}$ improvement from 42.16\% to 43.73\% through the use of the GRAZPEDWRI-DX dataset and data augmentation. Hardala \textit{et al.} \cite{hardalac2022wrist} developed an ensemble-based system for wrist fracture detection, attaining an AP$_{50}$ of 0.8639 on the Gazi University Hospital dataset. Yao \textit{et al.} \cite{yao2021rib} proposed a rib fracture detection system for chest CT scans, achieving an F1-score of 0.890, and demonstrated that AI-assisted diagnosis improved radiologists' recall and reduced diagnosis time by an average of 65.3 seconds. 

Attention mechanisms have been widely adopted to enhance feature selection. Hu \textit{et al.} \cite{hu2018squeeze} proposed the Squeeze-and-Excitation module, improves channel-wise feature calibration. Woo \textit{et al.} \cite{woo2018cbam} proposed the Convolutional Block Attention Module, which integrates channel and spatial attention to enhance performance on various tasks. Wan \textit{et al.} \cite{wan2023mixed} introduced the Mixed Layered Channel Attention mechanism, which refines texture information and outperforms SE by 1.0\% in mAP on Pascal VOC.

Chen \textit{et al.} \cite{chen2024mfmam} introduced a multi-scale feature fusion with an enhanced attention module for image inpainting, achieving superior PSNR, SSIM, and FID scores. Zhu \textit{et al.} \cite{zhu2024yolox} integrated a lightweight attention module into YOLOx for weed detection, improving AP by 1.16\% with 94.86\% accuracy in maize seedling fields. Lau \textit{et al.} \cite{lau2024lska} introduced Large Separable Kernel Attention, which reduces computational cost while maintaining competitive performance with models like LKA, ViTs \cite{liu2021Swin}, and ConvNeXt \cite{liu2022convnet}.

Despite these advancements, optimizing attention mechanisms for multi-scale feature extraction, computational efficiency, and generalization remains a challenge. YOLOv8 \cite{yolov8_ultralytics}, as the foundational model for numerous different domains work, balances speed, accuracy, and robustness, making it ideal for medical imaging tasks such as pediatric fracture detection.

\section{Method} 
\subsection{Motivation and Model}
\begin{figure}[h]
	\centering
	\includegraphics[width=0.9\linewidth]{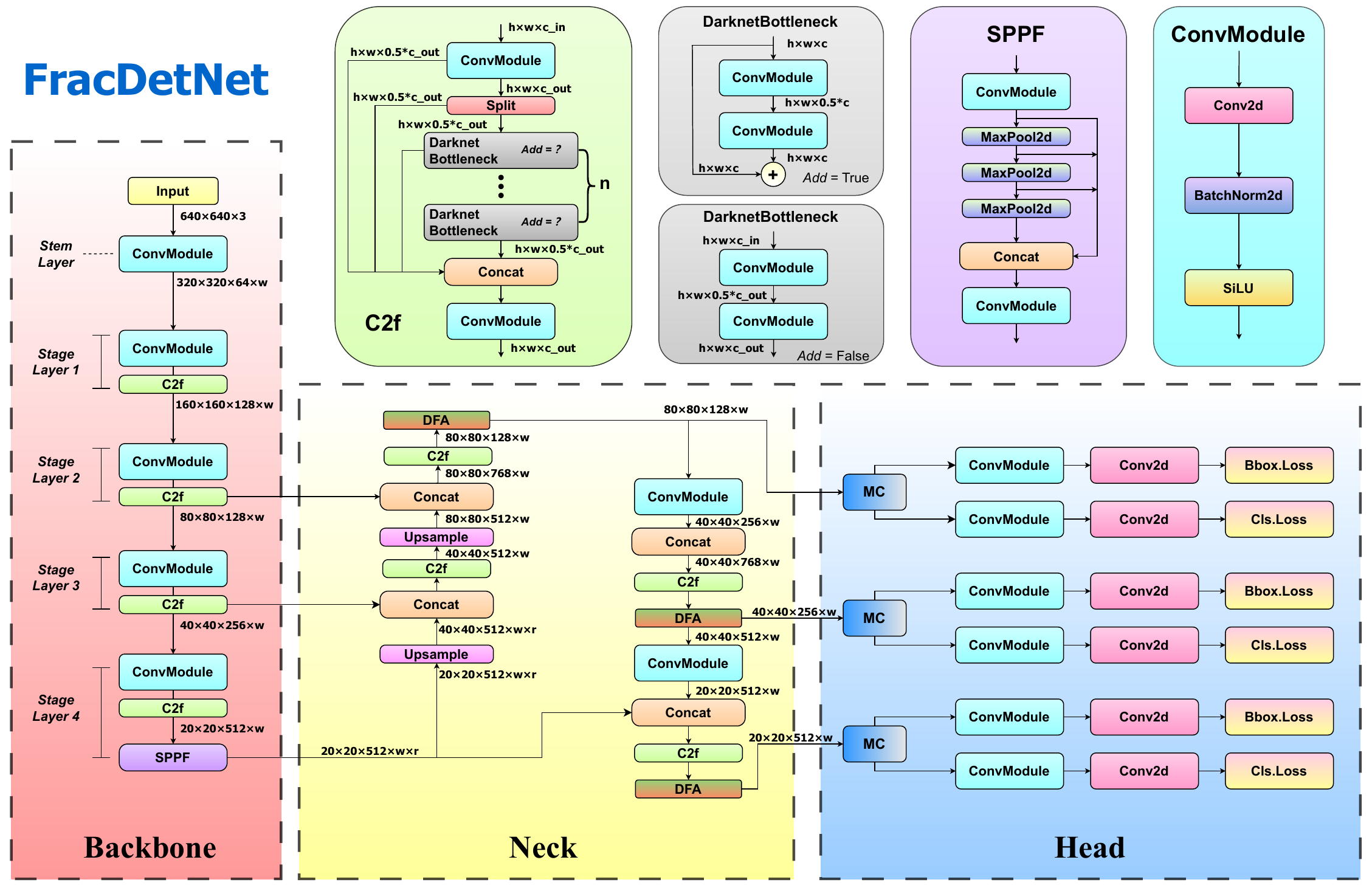}
	\caption{The overall architecture of FracDetNet, where the DFA and MC modules are improvements to the original architecture.}
	\label{fig2}
\end{figure}

Traditional fracture detection methods often face performance degradation when dealing with fractures of varying sizes and scales, particularly in medical imaging, where perspective effects and noise complicate the detection of small objects or complex shapes. To overcome these challenges, an enhanced FracDetNet framework is proposed, integrating self-attention and multi-scale deep convolution techniques to enhance detection accuracy and robustness. 

Our \textbf{enhanced FracDetNet} comprises the backbone network, the detection neck, and the detection head. (as shown in Fig. \ref{fig2}).    
\begin{itemize}
    \item 
The \textbf{backbone network} uses CSPDarknet, which reduces computational cost and improves feature extraction efficiency through a branching mechanism. By optimizing the network design, CSPDarknet achieves a substantial reduction in computational complexity without compromising precision. To further enhance inference speed, our FracDetNet incorporates an SPPF layer (Spatial Pyramid Pooling Fast) improving adaptability to objects of various sizes via multi-scale pooling operations.
   
       \item 
The \textbf{detection neck} is based on PAFPN \cite{zou2024detection}, which employs a bidirectional feature propagation mechanism to strengthen the representation of features across multiple scales \cite{joshi2020survey}, Addressing the shortcomings of conventional approaches in multi-scale object detection. PAFPN facilitates information flow between feature maps of different scales, improving localization and recognition of fractures. In medical imaging, perspective effects make it difficult for traditional methods to fuse global and local information effectively. Therefore, FracDetNet employs the Dual Focus Attention mechanism (DFA), which dynamically fuses local and global information, further enhancing feature map expressiveness and detection accuracy.
       
          \item    
Maintaining the lightweight characteristics of YOLO, The \textbf{detection head} outputs both bounding box coordinates and class labels directly from the extracted feature maps \cite{yolov8_ultralytics}. To address scale variation in fracture detection, the Multiscale Calibration Head (MC) is introduced, dynamically adjusting channel weights and applying multi-scale convolutions to improve feature extraction and fusion. This enables precise fracture localization across different scales.
\end{itemize}
\begin{figure}[h]
	\centering
	\includegraphics[width=1.0\linewidth]{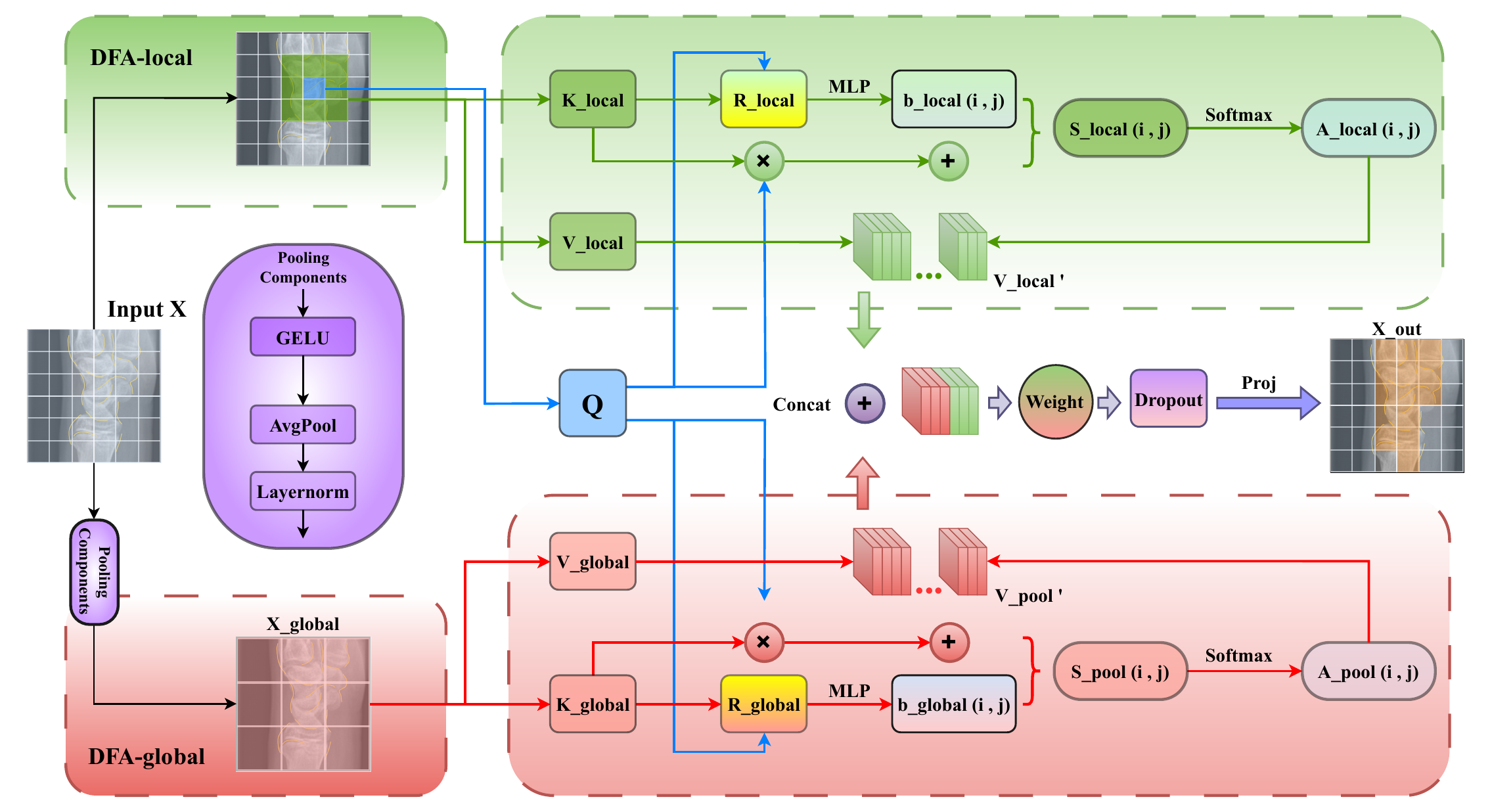}
	\caption{Overview of the proposed DFA. DFA is mainly computed in parallel by two work paths, DFA-local and DFA-global.}
	\label{fig3}
\end{figure}
\subsection{Dual-Focus Attention}
A major obstacle in the field of medical image analysis for fracture detection is how to identify fracture regions affected by varying imaging angles (i.e., the impact of perspective effects) when the image quality lower \cite{langerhuizen2019applications}. Due to the significant morphological differences among different types of fractures, the feature extraction process often captures a large amount of redundant information, which leads  to incomplete or ambiguous feature representations and  affects the detection. 

To better address  this issue, a Dual-Focus Attention (DFA) module is proposed in this work. It integrates local self-attention and global self-attention in the detection network. The local self-attention primarily captures fine-grained local structures and short-range dependencies, while the global self-attention focuses on extracting coarse-grained global semantics and long-range dependencies. 
Through this novel fusion, the detection network  can simultaneously address both the detailed features and the overall structure of the target, which achieves a more comprehensive feature representation.

For clarity, all bolded letters are the tensor representations for images and feature maps.  
Specifically, denote the input of DFA module as  feature map $\mathbf{X}$, which is  extracted by the backbone network and need to further process by PAFPN. 
The feature map $\mathbf{X}$ will be integrated into the two self-attention mechanisms: DFA-local and DFA-global  in parallel, where 
DFA-local aims to capture the local structural information of fracture regions and DFA-global is designed to acquire global information and long-range dependencies. 

Fig. \ref{fig3} shows the flow of the DFA module.  DFA consists of two parallel work paths, DFA-local and DFA-global. To compute the $\mathbf{V}_{\text{global}}$ in the global attention branch, we first down-sample the input feature map $\mathbf{X}$ using a pooling component. Here, $\mathbf{V}_{\text{global}}$ represents the global feature map computed from the global attention branch and $\mathbf{X}$ is the feature map input used for subsequent processing within the network. This operation can be formulated as follows: 
\begin{equation}
	\mathbf{X}_{\text{global}} = \mathit{LayerNorm} \bigl( \mathit{AvgPool} \bigl( \mathit{GELU} \bigl( \mathit{Linear} (\mathbf{X}) \bigr) \bigr) \bigr), 
\end{equation}
where \textit{Linear} transformation maps $\mathbf{X}$ to a different feature dimension for compression. The \textit{GELU} activation adds a smooth nonlinear transformation. The \textit{AvgPool} reduces the spatial resolution while preserving global information, and the \textit{LayerNorm} enhances numerical stability and accelerates convergence.  

Based on the feature maps are obtained above, some linear transformations are applied to the $\mathbf{X}_{\text{local}}$ and $\mathbf{X}_{\text{global}}$ (i.e., the inputs of the local and global branches) to generate the respective query ($Q$), key ($K$), and value ($V$) \cite{vaswani2017attention}, which we denoted as $\mathbf{Q}$, $\mathbf{K}_{\text{local}}$, $\mathbf{V}_{\text{local}}$, $\mathbf{K}_{\text{global}}$, and $\mathbf{V}_{\text{global}}$.

Meanwhile, to introduce learnable relative positional bias in the attention computation, a new relative position bias $\mathbf{R}(i,j)$ is established based on the positional differences between the query and the key. For instance, considering the local branch, the local relative position bias $\mathbf{R}_{\text{local}}(i,j)$ is given as follows:
\begin{equation}
	\mathbf{R}_{\text{local}}(i,j) = \mathit{Norm} \left( \left[ \frac{i - k}{H - 1}, \frac{j - l}{W - 1} \right] \right).
\end{equation} 
Here, $H$ and $W$ correspond to the height and width of the input feature map, respectively. $(i,j)$ denotes the query position and $(k,l)$ denotes the key position. The function $\mathit{Norm}(\cdot)$ is applied to normalize the relative position vector to facilitate subsequent computations.

According the $\mathbf{R}_{\text{local}}$ and $\mathbf{R}_{\text{global}}$, we can  compute the relative position bias matrices $\mathbf{b}_{\text{local}}$ and $\mathbf{b}_{\text{global}}$ in the Multi-Layer Perceptron (MLP) process.  
Specifically, for each relative position $\mathbf{R}_{\text{local}}(i,j)$ and $\mathbf{R}_{\text{global}}(i,j)$, the transformation through the MLP is formulated as follows:
\begin{equation}
	\mathbf{b}_{\text{local}}(i,j)
	= \mathrm{MLP}\bigl(\mathbf{R}(i,j)\bigr)
	= \mathbf{W}_2 \cdot \bigl(\mathit{ReLU} \bigl(\mathbf{W}_1 \cdot \mathbf{R}_{\text{local}}(i,j) + \mathit{b}_1 \bigr)\bigr) + \mathit{b}_2.
\end{equation}
Here, we take the local branch as an example and the $\mathbf{W}_1$ and $\mathbf{W}_2$ are trainable weight matrices, while $\mathit{b}_1$ and $\mathit{b}_2$ represent learnable bias terms. The function $\textit{ReLU}(\cdot)$ denotes the nonlinear activation function ReLU \cite{glorot2011deep}. This process effectively encodes relative positional information, strengthening the model’s capability to capture spatial relationships across the feature space. 

Next, both the query $\mathbf{Q}$ and key $\mathbf{K}$ are normalized by their $L_2$ norms:
\begin{equation}
	\mathbf{\widehat{Q}} = \frac{\mathbf{Q}}{\|\mathbf{Q}\|_2}
	\quad\text{and}\quad
	\mathbf{\widehat{K}} = \frac{\mathbf{K}}{\|\mathbf{K}\|_2}.
\end{equation} 
Here, L2 regularization is a technique that introduces a regularization term into the loss function that is proportional to the squared sum of the model’s weights, discouraging large weights and reducing overfitting \cite{hoerl1970ridge}.

Then we compute two similarity matrices $\mathbf{S}_{\text{local}}$ and $\mathbf{S}_{\text{global}}$ as: 
\begin{equation}
	\mathbf{S}_{\text{local}}(i,j)
	= \mathbf{\widehat{Q}}(i,j)\cdot \mathbf{\widehat{K}}_{\text{local}}(i,j)^T
	+ \mathbf{b}_{\text{local}}(i,j),
\end{equation}
\begin{equation}
	\mathbf{S}_{\text{global}}(i,j)
	= \mathbf{\widehat{Q}}(i,j)\cdot \mathbf{\widehat{K}}_{\text{global}}(i,j)^T
	+ \mathbf{b}_{\text{global}}(i,j).
\end{equation}
Here, we denote  $\mathbf{\widehat{Q}}(i,j)$ and $\mathbf{\widehat{K}}(i,j)$ as the query and key vectors after $L_2$ normalization. The  $\mathbf{b}_{\text{local}}(i,j)$ and $\mathbf{b}_{\text{local}}(i,j)$ are the learnable bias generated based on the relative positional coordinates.   
Subsequently, the Softmax function is applied to each elements in the similarity matrix $\mathbf{S}(i,j)$ to obtain the attention weights $\mathbf{A}(i,j)$.  The computation for the local and global branches attentions are formulated as follows:
\begin{equation}
	\mathbf{A}_{\text{local}}(i,j) = \mathit{Softmax}(\mathbf{S}_{\text{local}}(i,j)), \quad
	\mathbf{A}_{\text{global}}(i,j) = \mathit{Softmax}(\mathbf{S}_{\text{global}}(i,j)).
\end{equation}
Then we can multiply them by the corresponding value $\mathbf{V}$ to compute the final attention output as: 
\begin{equation}
	\mathbf{V}'_{\text{local}} = \mathbf{A}_{\text{local}} \cdot \mathbf{V}_{\text{local}}, \quad
	\mathbf{V}'_{\text{global}} = \mathbf{A}_{\text{global}} \cdot \mathbf{V}_{\text{global}}.
\end{equation}
After obtaining $\mathbf{V}'_{\text{local}}$ and $\mathbf{V}'_{\text{global}}$, these two feature components are concatenated and then fused through a weighted combination. The fused representation is further processed via a \textit{Dropout} and projection layer (\textit{proj}) to yield the ultimate output feature representation as:
\begin{equation}
	\mathbf{X}_{\text{out}} = \mathbf{X}+\mathit{Dropout}(\mathit{proj}(\mathit{concat}(\mathbf{V'}_{\text{local}}, \mathbf{V'}_{\text{global}}))).
\end{equation}

\begin{figure}[h]
	\centering
	\includegraphics[width=0.9\linewidth]{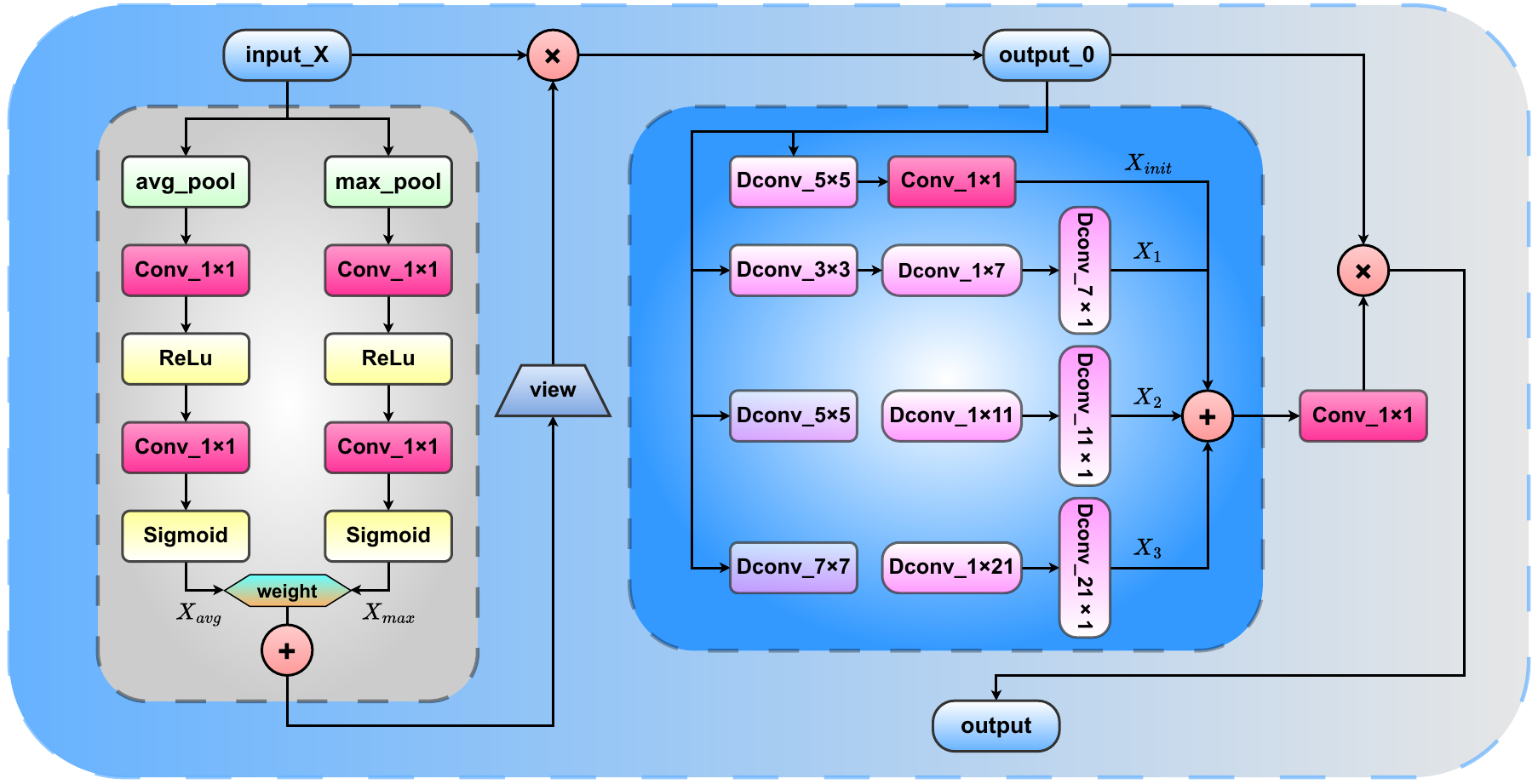}
	\caption{Overview of the proposed MC. MC is mainly computed in parallel by two parts, Channel Block and Multi-scale Block.}
	\label{fig4}
\end{figure}

\subsection{Multi-scale Calibration Head}
The multi-scale information of fracture regions presents challenges to the precision of detection during inference. Fine-grained crack identification relies on high-frequency details, while large-scale fractures require global structural and morphological features. 
To address this issue, a Multi-scale Calibration head (MC) is integrated into the detection head, with a dual-path feature processing mechanism utilized to optimize multi-dimensional feature representation. 

This module first applies a channel-wise dynamic weighting mechanism, which adaptively reweight  feature channels via learnable parameters to enhance critical semantic information. 
Depthwise separable convolutions are then utilized to form parallel multi-scale feature extraction branches, which captures spatial features at different receptive fields. This design enables hierarchical fusion and adaptive enhancement of multi-scale features, which enables the model to concentrate on fracture texture and structural deformation, thereby improving detection robustness. 

The structure of the MC is shown in Fig. \ref{fig4}. Given an input $\mathbf{X}$, the MC initially employs global average pooling and global max pooling to independently compute two channel-attention sub-branches. The computation processes are as follows: 
\begin{equation}
	\mathbf{X}_{\text{avg}} = \sigma \left( \mathit{Conv}_{1\times1} \left( \mathit{ReLU} \left( \mathit{Conv}_{1\times1} \left( \mathit{Avgpool} (\mathbf{X}) \right) \right) \right) \right),
\end{equation}
\begin{equation}
\mathbf{X}_{\text{max}} = \sigma \left( \mathit{Conv}_{1\times1} \left( \mathit{ReLU} \left( \mathit{Conv}_{1\times1} \left( \mathit{Maxpool} (\mathbf{X}) \right) \right) \right) \right).
\end{equation}
Here \(\mathbf{X}\) denotes the input feature map, \(\mathit{Avgpool} (\mathbf{X})\) and \(\mathit{Maxpool} (\mathbf{X})\) correspond to the global average pooling and the global max pooling operations, respectively. 
\(\mathit{Conv}_{1\times1} (\cdot)\) denotes a \(1 \times 1\) convolution for feature refinement, while \(\mathit{ReLU} (\cdot)\) introduces non-linearity. 
The sigmoid function \(\sigma (\cdot)\) scales the output to \([0,1]\) for attention weighting. 
The   \(\mathbf{X}_{\text{max}}\) and 
\(\mathbf{X}_{\text{avg}}\) serve as attention weights to enhance feature representation. 

After obtaining \( \mathbf{X}_{\text{avg}} \) and \( \mathbf{X}_{\text{max}} \), they are combined through weighted summation. 
Subsequently, a \textit{view} operation is applied to adjust the dimension before weighting the original feature map, i.e., 
\begin{equation}
	\textbf{X}_{output0} = \mathbf{X} \cdot \mathit{view}(\mathbf{X}_{\text{max}} + \mathbf{X}_{\text{avg}}), 
\end{equation}
where \(\mathit{view}\) reshapes the attention weights to match the dimensions of \(\mathbf{X}\)  and \(\textbf{X}_{output0}\) is the final weighted feature map obtained by element-wise multiplication with the original feature map \(\mathbf{X}\). 

To balance model efficiency and performance, a depthwise convolution is introduced \cite{huang2024channel}. 
It applies independent spatial kernels to each input channel, which reduces parameter sizes and computational complexity. To enhance multi-scale fracture feature extraction, a hierarchical learning framework is used: first, the depthwise convolution captures the mid-scale spatial context from channel-weighted feature maps. Then, a $(1 \times 1)$ convolution refines the feature representation while preserving original information for stable fusion. 
In fact, directly applying a large kernel (e.g., $(n \times n)$) may excessively smooth features, which leads to the loss of high-frequency details. To mitigate this issue, a progressive refinement strategy is employed, where small-kernel convolutions are first used to extract local features before the large-kernel depthwise convolution.  

To alleviate the computational bottleneck of large-kernel convolution, an axial decomposition convolution approach is adopted. It decomposes  two-dimensional large-kernel convolution into two consecutive one-dimensional operations: first, a $(n \times 1)$ depthwise convolution captures horizontal spatial features, followed by a $(1 \times n)$ depthwise convolution to extract vertical correlations. The specific computation process is as follows:  
\begin{equation}
	\left\{
	\begin{aligned}
		\mathbf{X}_{\text{init}} &= DW_{5 \times 5} \left( Conv_{1 \times 1} (\textbf{X}_{output0})  \right), \\
		\mathbf{X}_1 &= DW_{7 \times 1} \left( DW_{1 \times 7} \left( DW_{3 \times 3} (\textbf{X}_{output0}) \right) \right), \\
		\mathbf{X}_2 &= DW_{11 \times 1} \left( DW_{1 \times 11} \left( DW_{5 \times 5} (\textbf{X}_{output0}) \right) \right), \\
		\mathbf{X}_3 &= DW_{21 \times 1} \left( DW_{1 \times 21} \left( DW_{7 \times 7} (\textbf{X}_{output0}) \right) \right),
	\end{aligned}
	\right.
\end{equation}
where \( DW_{k \times n} (\cdot) \) represents a depthwise convolution with a kernel size of \( k \times n \).
The axial decomposition strategy maintains an equivalent \( n \times n \) receptive field while decreasing the computational burden from \( o(k^2C) \) to \( o(2kC) \), where \( k \) represents the kernel size and \( C \) denotes the number of channels. The method maintains an equilibrium between computational performance and functional effectiveness to represent features. 

Building on multi-scale features, feature fusion is carried out to combine information from diverse receptive fields. 
Subsequently, a (\( 1 \times 1 \)) convolution is combined features for additional processing, enhancing feature representation and generating a spatial attention map. 
Finally, this spatial attention map is used to weight \( \textbf{X}_{\text{output0}} \), which enhances the response of critical regions. The specific computation process is as follows:
\begin{equation}
	\mathbf{X}_{\text{final}} = \mathit{Conv}_{1 \times 1} \left( \textbf{X}_{output0} \otimes \left( \mathit{Conv}_{1 \times 1} \left( \mathbf{X}_{\text{init}} + \mathbf{X}_1 + \mathbf{X}_2 + \mathbf{X}_3 \right) \right)  \right), 
\end{equation}
where \(\mathbf{X}_{\text{final}}\) represents the final output feature map, \(\mathbf{X}_{\text{init}}\) is the initial input feature map, and \(\mathbf{X}_1, \mathbf{X}_2, \mathbf{X}_3\) denote the multi-scale feature representations extracted using different convolutional kernels. \(\mathit{Conv}_{1 \times 1} (\cdot)\) represents a \(1 \times 1\) convolution operation, which is used for feature transformation and refinement. \(\otimes\) represents Hadamard product, which applies the attention weights to enhance or suppress specific spatial locations in the feature map.

\section{Experiment}
\subsection{Experiment Setting} 

\textbf{Environment: } 
All experiments are performed on a single NVIDIA RTX 3090 GPU using FracDetNet and other object detectors implemented with MMYOLO \cite{mmyolo2022} and MMDetection \cite{mmdetection}. 
The environment includes Python 3.8, CUDA 11.8, PyTorch 2.0.0, and Torchvision 0.15.1. 
The initial learning rate is set to 0.01, with \(640 \times 640\) input images and the batch size  $16$. Models are initialized with pre-trained weights from the COCO dataset to improve training efficiency.

\textbf{Dataset: } 
We use the GRAZPEDWRI-DX dataset \cite{nagy2022grazpedwri}, which is provided by the Medical University of Graz. It includes 20,327 pediatric wrist trauma X-ray images from 6,091 patients, spanning from 2008 to 2018. It includes 74,459 image-level labels and 67,771 annotated targets, covering various fracture types.

\textbf{Comparison: } 
The comparison methods we used include 
the Vfnet (ResNet-0)~\cite{zhang2020varifocalnet}, Retinanet (ResNet-50) \cite{lin2017focal}, 
Faster R-CNN (ResNet-50) \cite{Ren_2017}, 
Cascade R-CNN (ResNet-50) \cite{Cai_2019}, 
Mask R-CNN (Swin-T)~\cite{liu2021Swin}, 
Mask R-CNN (ConvNext), Dino 4-scale R-50~\cite{zhang2022dino}, YOLOv5s~\cite{glenn_jocher_2022_7002879}, YOLOv7tiny~\cite{wang2022yolov7}, YOLOXs~\cite{yolox2021}, 
YOLOv8 \cite{ju2023fracture}, 
and YOLOv8-ResCBAM \cite{ju2024yolov8}.

\textbf{Evaluation: }  
Mean Average Precision (mAP) is used as the evaluation metric, with mAP calculated over IoU thresholds from $0.5$ to $0.95$, which emphasizes  \textit{mAP$_{50}$} at IoU 0.5. Results for small, medium, large, and fracture-specific targets are  reported as \textit{mAP$_S$}, \textit{mAP$_M$}, \textit{mAP$_L$}, and \textit{mAP$_{Fracture}$}. Additionally, floating-point operations (FLOPs) are  computed to assess computational cost, and the model's parameter size is reported to evaluate complexity.
\begin{table}[t]
	\centering
	\caption{The performance comparison on GRAZPEDWRI-DX, where the bolded values indicate the best-performing method under each metric.}
	\label{tab:2}
	\resizebox{\textwidth}{!}{ 
	\fontsize{10}{10}\selectfont 
		\begin{tabular}{lcccccccc}
			\toprule
			\textbf{Methods} & \textbf{mAP(\%)} & \textbf{mAP$_{50}$(\%)} & \textbf{mAP$_S$(\%)} & \textbf{mAP$_M$(\%)} & \textbf{mAP$_L$(\%)} & \textbf{mAP$_{Fracture}$(\%)} & \textbf{Flops(T)} & \textbf{Parameters(M)}\\
			\midrule
			Vfnet~\cite{zhang2020varifocalnet} & 32.5 & 55.7 & 16.6 & 30.2 & 35.1 & 52.3 & 0.047 & 32.90 \\
			Retinanet~\cite{lin2017focal} & 25.1 & 46.5 & 12.0 & 21.3 & 27.7 & 50.4 & 0.050 & 36.52 \\
			Faster R-CNN~\cite{Ren_2017} & 28.6 & 53.7 & 12.6 & 20.7 & 31.7 & 49.7 & 0.061 & 32.89 \\
			Cascade R-CNN~\cite{Cai_2019} & 19.6 & 34.3 & 11.7 & 25.7 & 22.6 & 49.3 & 0.088 & 69.18 \\
			Mask R-CNN (Swin-T)~\cite{liu2021Swin} & 34.0 & 58.9  & 14.6 & 25.0 & 36.4 & 50.6 & 0.244 & 47.79 \\
			Mask R-CNN (Convnext)~\cite{liu2022convnet} & 37.2 & 62.5  & 19.0 & 35.5 & 37.9 & 51.6 & 0.238 &  47.72\\
			Dino 4-scale R-50~\cite{zhang2022dino}  & 37.3 & 61.0 & 14.4 & 33.2 & 40.4 & 54.6 & 0.245 & 47.70 \\
			\midrule
			YOLOv5s~\cite{glenn_jocher_2022_7002879}  & 25.4 & 43.8 & 13.0 & 24.1 & 28.3 & 49.7 & 0.008 & 7.05\\
			YOLOv7tiny~\cite{wang2022yolov7}  & 31.7 & 55.1 & 14.0 & 29.3 & 33.7 & 51.8 & \textbf{0.007} & \textbf{6.04} \\
			YOLOXs~\cite{yolox2021} & 36.9 & 63.9 & \textbf{17.8} & 31.9 & 38.3 & 53.8 & 0.013 & 8.94 \\
			YOLOv8s~\cite{yolov8_ultralytics} (Baseline) & 37.2 & 61.3 & 10.4 & 26.9 & 40.6 & 54.5 & 0.014 & 22.97 \\
			\midrule
			YOLOv8\cite{ju2023fracture} & 39.2 & 61.2 & - & - & - & - & 0.029 & 11.13 \\
			YOLOv8-AM (ResCBAM)\cite{ju2024yolov8} & 38.9 & 61.6 & - & - & - & - & 0.038 & 16.06 \\
			YOLOv8-AM (SA)\cite{ju2024yolov8} & 39.0 & 62.7 & - & - & - & - & 0.028 & 11.14 \\
			YOLOv8-AM (ECA)\cite{ju2024yolov8} & 37.4 & 61.4 & - & - & - & - & 0.028 & 11.14 \\
			YOLOv8-AM (GAM)\cite{ju2024yolov8} & 39.7 & 62.5 & - & - & - & - & 0.034 & 13.86 \\
			YOLOv8-AM (ResGAM)\cite{ju2024yolov8} & 38.6 & 61.4 & - & - & - & - & 0.034 & 13.86 \\
			\textbf{Ours} & \textbf{40.0} & \textbf{63.9} & 13.5 & \textbf{35.5} & \textbf{41.9} & \textbf{56.1} &0.016 &24.96 \\
			\bottomrule
		\end{tabular}
	}
\end{table}

\subsection{Comparison Results} 
Table \ref{tab:2} shows the comparison of the method we proposed with several state-of-the-art detectors. We can find that our method demonstrates superior performance, achieving $40.0\%$ in \textit{mAP}, $63.9\%$ in \textit{mAP$_{50}$}, and $56.1\%$ in \textit{mAP$_{Fracture}$}, outperforming all existing approaches. Compared to YOLOv8s (baseline), our method improves by \textbf{$7.5\%$} in \textit{mAP$_{50-95}$}, \textbf{$4.2\%$} in \textit{mAP$_{50}$}, and \textbf{$2.9\%$} in \textit{mAP$_{Fracture}$}. The improvement in performance is primarily attributed to a more focused attention on features and an enhanced capability for multi-scale object detection.

Furthermore, our method demonstrates significant advantages in detecting objects of different scales, achieving 35.5\% in medium-sized object detection (\textit{mAP$_M$}) and 41.9\% in large-sized object detection (\textit{mAP$_L$}), significantly surpassing the baseline methods. Notably, compared to YOLOXs, our method improves \textit{mAP} by \textbf{8.4\%} while also achieving superior performance in \textit{mAP$_M$} and \textit{mAP$_L$}. However, in terms of computational efficiency, our method incurs an increase in FLOPs and parameter size. Nevertheless, the notable enhancement in detection accuracy demonstrates the validity of our method, positioning it as a competitive choice for object detection applications.
\begin{table*}[t]
	\centering
	\caption{The performance of our method across different models, assessing the contribution of each component to the overall model.}
	\label{tab:4}
	\setlength{\tabcolsep}{3pt} 
	\fontsize{6}{4}\selectfont 
		\begin{tabular}{c c c c c c c}
			\toprule
			\textbf{Methods} & \textbf{DFA} & \textbf{MC}  & \textbf{mAP(\%)} & \textbf{mAP$_{50}$(\%)} & \textbf{Flops(T)} & \textbf{Parameters(M)} \\
			\midrule
			\multirow{4}{*}{YOLOv5s}       &                     &               & 25.4               & 43.8                     & 0.008               & 7.05 \\
			& \checkmark         &               & 28.7               & 49.3                     &     0.008          & 8.35 \\
			&                    & \checkmark   & 29.3      & 49.8            &        0.009       & 7.74 \\
			& \checkmark         & \checkmark   & \textbf{29.8}      & \textbf{50.9}            &      0.010         & 9.04 \\
			\midrule
			\multirow{4}{*}{YOLOXs}        &                     &               & 36.9               & 63.9                     & 0.013              & 8.94 \\
			& \checkmark         &               & 38.6               & \textbf{64.9}          &     0.014          & 10.24 \\
			&                    & \checkmark   &   38.3    &   64.3    &     0.014          & 9.63 \\
			& \checkmark         & \checkmark   &  \textbf{38.8}     &   64.2    &      0.016         &  10.93\\
			\midrule
			\multirow{4}{*}{YOLOv7tiny}       &                &             & 31.7        & 55.1           &   0.007     &  6.04\\
			& \checkmark         &               & 32.5               & 56.8                     &   0.007     & 7.34 \\
			&                    & \checkmark   & 33.9      & 57.2            &        0.008       & 6.73 \\
			& \checkmark         & \checkmark   & \textbf{34.8}      & \textbf{59.5}            &     0.009     & 8.03 \\
			\midrule
			\multirow{4}{*}{YOLOv8s}       &                     &               & 37.2               & 61.3                     & 0.014              & 22.97 \\
			& \checkmark         &               & 38.9               & 63.8                     &     0.014     & 24.27 \\
			&                    & \checkmark   & 39.5      & 63.2            &      0.015         &  23.66\\
			& \checkmark         & \checkmark   & \textbf{40.0}      & \textbf{63.8}            &      0.016    & 24.96 \\
			\bottomrule
		\end{tabular}
	
\end{table*}

\subsection{Ablation Results}
To further evaluate the effectiveness of the proposed method, the DFA and MC modules were integrated into multiple object detection models, including YOLOv5s, YOLOXs, YOLOv7tiny, and YOLOv8s. Their impact on mAP and computational cost was analyzed to further verify their generalization capability. As shown in Table \ref{tab:4}, each module individually enhances the mAP of the baseline model, while the optimal performance is achieved when both modules are combined. For YOLOv8s, the combination of both modules leads to a 3.9\% improvement in \textit{mAP$_{50}$}. The experimental findings suggest that, despite the increased model complexity and parameter count introduced by these modules, all detection models exhibit a significant improvement in mAP, providing additional evidence of the proposed method’s effectiveness.
\begin{figure}[h]
	\centering
	\includegraphics[width=1\columnwidth]{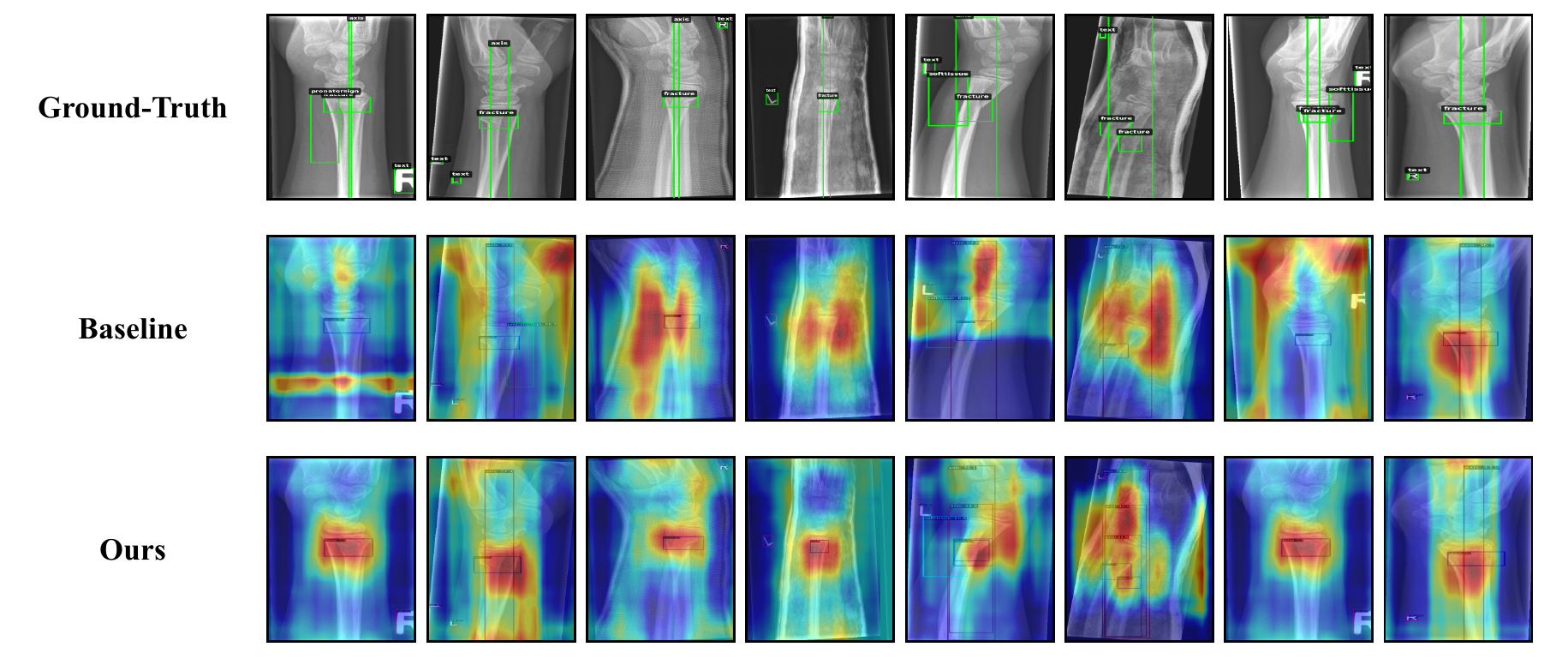}
	\caption{The detection results between the baseline model (YOLOv8s) and FracDetNet. The top row shows the original fracture images with ground-truth, where the red boxes represent detected fracture boxes. The middle row presents the heatmaps generated by the baseline model, and the bottom row displays the heatmaps generated by our method.}
	\label{fig5}
\end{figure}

\subsection{Visualization}
To further validate the effectiveness of FracDetNet from a visualization perspective, Grad-CAM \cite{mmyolo2022} is employed to visualize the final output feature maps of both the baseline model and FracDetNet. As shown in the Fig.~\ref{fig5}, compare  to the baseline model, which fails to focus on certain fracture regions, our approach intuitively concentrates on the fracture areas. Additionally, for fracture regions that the baseline model can identify, our method achieves a more precise focus on the fracture locations.

\section{Conclusion}
This paper proposes an enhanced Fracture Detection Network (\text{FracDetNet}), integrating Dual-Focus Attention (DFA) and Multi-scale Calibration Head (MC) to address challenges in detecting subtle and diverse fractures in medical X-ray imaging. DFA captures global and local information, while MC enhances multi-scale feature aggregation, ensuring robustness in complex clinical scenarios. Experimental results demonstrate that \text{FracDetNet} considerably outperforms the baseline model across multiple evaluation metrics.
Compared to the baseline, \text{FracDetNet} improves by $4.2\%$ in $\textit{mAP}_{50}$ and $7.5\%$ in $\textit{mAP}_{50\text{-}95}$, while maintaining competitive computational efficiency. Our approach also excels in detection accuracy, robustness, generalization, and clinical applicability. Further analysis demonstrates its strong adaptability across varying object scales and imaging conditions, highlighting its potential in pediatric fracture detection.

\section*{Acknowledgements}
This work was supported by the National Natural Science Foundation of China under Grant Nos. 62276111. It was also partly supported by the Grants of Huazhong Agricultural University under Grant No. 2662024XXPY005.
%
%
%
\bibliographystyle{splncs04}
\bibliography{mybibliography}

\end{document}